\documentclass[letterpaper]{article} 
\usepackage{aaai2026}  
\usepackage{times}  
\usepackage{helvet}  
\usepackage{courier}  
\usepackage[hyphens]{url}  
\usepackage{graphicx} 
\usepackage[numbers]{natbib}
\usepackage{caption} 
\usepackage{algorithm,algpseudocode}
\usepackage{amsfonts}
\usepackage{amsmath}
\usepackage{multirow}
\usepackage[most]{tcolorbox}
\usepackage{xcolor}
\usepackage{enumitem}
\usepackage{newfloat}
\usepackage{listings}
\DeclareCaptionStyle{ruled}{labelfont=normalfont,labelsep=colon,strut=off} 
\floatstyle{ruled}
\newfloat{listing}{tb}{lst}{}
\floatname{listing}{Listing}
\nocopyright
\usepackage{cleveref}

\title{Patho-AgenticRAG: Towards Multimodal Agentic Retrieval-Augmented Generation for Pathology VLMs via Reinforcement Learning}
\author{
    \textbf{Wenchuan Zhang}$^{1,2}$\thanks{Equal contribution.} \quad
\textbf{Jingru Guo}$^{3}$\footnotemark[1] \quad
\textbf{Hengzhe Zhang}$^{4}$\footnotemark[1] \quad
\textbf{Penghao Zhang}$^{5}$ \quad
\textbf{Jie Chen}$^{2}$ \quad \\
\textbf{Shuwan Zhang}$^{6}$ \quad 
\textbf{Zhang Zhang}$^{1}$ \quad
\textbf{Yuhao Yi}$^{1,2}$\thanks{Corresponding author.} \quad
\textbf{Hong Bu}$^{1,2}$
}
\affiliations{
    \textsuperscript{\rm 1}Department of Pathology, West China Hospital, Sichuan University\\
    \textsuperscript{\rm 2}Institute of Clinical Pathology, West China Hospital, Sichuan University\\
    \textsuperscript{\rm 3}University of Toronto 
    \textsuperscript{\rm 4}School of Engineering and Computer Science, Victoria University of Wellington\\
    \textsuperscript{\rm 5}Independent Researcher
    \textsuperscript{\rm 6}Department of Pathology, Shengjing Hospital of China Medical University\\
    \texttt{zhangwenchuan@stu.scu.edu.cn, yuhaoyi@scu.edu.cn}
}

\usepackage{bibentry}

\begin{document}

\maketitle

\begin{abstract}
Although Vision Language Models (VLMs) have shown strong generalization in medical imaging, pathology presents unique challenges due to ultra-high resolution, complex tissue structures, and nuanced clinical semantics. These factors make pathology VLMs prone to hallucinations, i.e., generating outputs inconsistent with visual evidence, which undermines clinical trust. Existing RAG approaches in this domain largely depend on text-based knowledge bases, limiting their ability to leverage diagnostic visual cues. To address this, we propose Patho-AgenticRAG, a multimodal RAG framework with a database built on page-level embeddings from authoritative pathology textbooks. Unlike traditional text-only retrieval systems, it supports joint text–image search, enabling direct retrieval of textbook pages that contain both the queried text and relevant visual cues, thus avoiding the loss of critical image-based information. Patho-AgenticRAG also supports reasoning, task decomposition, and multi-turn search interactions, improving accuracy in complex diagnostic scenarios. Experiments show that Patho-AgenticRAG significantly outperforms existing multimodal models in complex pathology tasks like multiple-choice diagnosis and visual question answering. Our project is available at the Patho-AgenticRAG repository: 
  \url{https://github.com/Wenchuan-Zhang/Patho-AgenticRAG}.
\end{abstract}

\begin{figure*}
    \centering
    \includegraphics[width=1\linewidth]{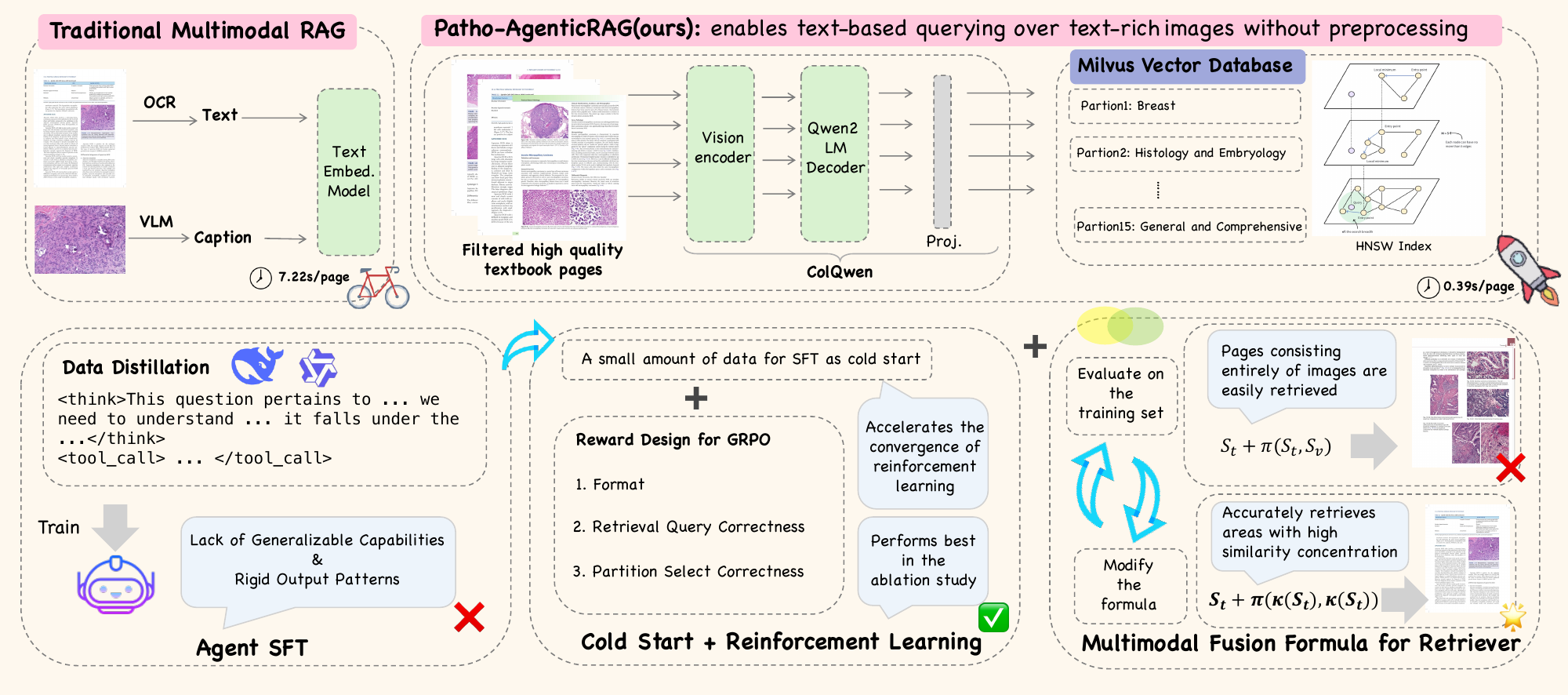}
    \caption{Knowledge Base Construction and Agent Training Method in the Patho-AgenticRAG}
    \label{fig:workflow}
\end{figure*}

\section{Introduction}
With the continuous development of large-scale vision-language models (VLMs), multimodal learning has made breakthrough progress in many fields such as natural image understanding, image-text generation, and medical image analysis. Compared with other medical images (such as X-rays, CT, and MRI), pathological images, with their ultra-high resolution, fine-grained structure, and complex semantic relationships, have put forward higher requirements on the perception, reasoning, and factual consistency capabilities of the model. 

In recent years, more and more studies have attempted to introduce VLMs into digital pathology tasks, such as diagnostic assistance \cite{PathChatplus}, risk stratification \cite{vlsa}, and question-answering systems \cite{pathchat}. However, existing pathology VLMs still face key challenges such as severe hallucinations and a lack of structured semantic control of retrieval mechanisms, especially in tasks that require factual support and traceable evidence. Therefore, how to build a highly reliable, multimodally interpretable pathology VLM with a factual consistency assurance mechanism has become an important issue that needs to be solved in this field. Although many works in recent years have attempted to apply the Retrieval-Augmented Generation (RAG) framework to medical multimodal tasks to improve the accuracy and credibility of large language models in medical reasoning, there are still many limitations in pathological image analysis scenarios. MMed-RAG proposed a general multimodal medical RAG system \cite{mmedrag}, but the image vector library it constructed lacks fine-grained annotation of the organization system and is only divided according to the type of imaging modality (such as CT, X-ray), which cannot meet the actual needs of ``classification of the organ or tissue system to which the image belongs" in pathological diagnosis. For example, in clinical work, it is necessary to clarify whether the image comes from a specific system, such as breast or lung, to match contextual knowledge and structured diagnostic paths. At the same time, some studies completely ignore the image modality and rely only on text retrieval, failing to give full play to the important role of images in assisting VLM reasoning, especially in scenarios facing image-text consistency and visual evidence support \cite{jabal2024language, cheetirala2025less}. In addition, recent methods such as MedRAG and Medical Graph RAG introduce complex reasoning processes guided by knowledge graphs \cite{medrag, medicalgraphrag}. Although they enhance reasoning capabilities, the system design is complicated, the execution process lacks intelligence and scalability, and it is difficult to adapt to a variety of actual clinical tasks. Although Liu et al. emphasizes the importance of knowledge augmentation, its RAG module has the problem of insufficient instruction following, making it difficult to stably extract relevant knowledge fragments in complex instruction following tasks, affecting model performance \cite{react}. This study aims to build an intelligent retrieval augmented generation framework for pathology VLMs, to improve the credibility and interpretability of the model in complex question-answering and reasoning tasks. 

Our method focuses on three dimensions: multimodal knowledge retrieval, intelligent planning capabilities, and adaptability to pathology scenarios. Different from the previous RAG framework based on static prompts or text retrieval, Patho-AgenticRAG introduces a multimodal image-text retrieval module and an agent mechanism with planning capabilities, which enables the model to more effectively retrieve target images and corresponding knowledge content from the structured pathology knowledge base, and to reasonably integrate and reason. In addition, we introduce a reinforcement learning optimization strategy to make the agent more robust and generalizable in the highly complex and uncertain question-answering environment in the field of pathology (see Figure~\ref{fig:workflow}). The main contributions are summarized as follows:
\begin{itemize}
\item We proposed \textbf{a novel Multimodal Retrieval Mechanism} that combines multimodal (image-text) vector space modeling with a tissue-aware retrieval strategy. This significantly improves the recall rate of the target knowledge fragment while ensuring accuracy, providing a guarantee for fine-grained knowledge alignment in pathology diagnosis tasks.

\item We built \textbf{a planning-capable intelligent agent within the Agentic RAG system}, which autonomously plans multi-round retrieval and reasoning trajectories in response to complex natural language pathology questions. It dynamically invokes relevant multimodal knowledge and effectively supports long-term dependency modeling and multi-hop reasoning in diagnostic tasks.

\item We proposed \textbf{a Tool-Integrated Reasoning training paradigm tailored for medical diagnostics}, built upon GRPO. This paradigm enables the agent to make fine-grained decisions, such as whether to invoke retrieval, how to reformulate questions, and how to assign domain-specific tools or classifiers within complex pathology question answering scenarios. It addresses the high-stakes nature of medical reasoning by promoting robust decision-making and reliable tool coordination.
\end{itemize}

\section{Related Work}
\subsection{Multimodal Agentic RAG}
Retrieval-Augmented Generation mitigates the limitations of large language models in knowledge completeness and factual consistency by incorporating external knowledge sources \cite{rag, llmhallucination}. Traditional RAG systems, such as Na\"ive RAG \cite{bgelargeenv15, Nvembed} and Advanced RAG \cite{visrag, m3docrag}, typically follow a static, linear ``retrieve-then-read" workflow. While effective for simple queries, they struggle with complex tasks requiring multi-step reasoning, context-aware adaptation, or tool use \cite{agenticragreview}.
Agentic RAG extends the RAG paradigm by embedding autonomous agents into the retrieval and reasoning process \cite{agenticrag}. These agents can plan retrieval strategies \cite{reaper}, invoke external tools \cite{vragrl}, reflect and revise outputs \cite{agenticragreflection}, and collaborate across multiple roles or modalities \cite{vidorag, searchr1, mmsearchr1}. Depending on design, they may operate as single decision-makers (e.g., router agents) or as specialized multi-agent systems for handling heterogeneous data sources. 
Despite these advances, existing Agentic RAG research in the medical domain has focused predominantly on text or structured data \cite{medicalagenticrag2text1, medicalagenticrag2text2}, overlooking the diagnostic richness encoded in medical images. However, fields like pathology rely heavily on visual features, patterns in tissue morphology, staining, and spatial arrangements that cannot be captured through text alone.

\subsection{Reinforcement Learning for Medical VLMs}

Reinforcement learning (RL) provides a promising paradigm for aligning the outputs of VLMs with clinical accuracy requirements, especially in high-risk medical domains where hallucinated descriptions can lead to severe consequences \cite{pathor1}. A core challenge in medical VLMs is ensuring factual alignment between visual evidence (e.g., pathology slides, radiology images) and textual outputs \cite{cpathagent, huatuogptvision,xuzhepathrl}. 

However, direct end-to-end RL of large VLMs remains highly impractical. The main limitations include the scarcity of high-quality reward data from physicians \cite{rarl}, the instability of RL training on large models, and the lack of interpretability in the learned behaviors \cite{zhu2025toward}. These issues make it difficult to deploy RL-fine-tuned models safely in clinical settings.
To address this, recent approaches have turned to agent-centric strategies, where RL is applied not to the internal parameters of VLMs, but to the decision-making policies of external agents that interact with them. These agents may learn how to craft better queries, validate initial answers, or select relevant external knowledge iteratively \cite{mmedrag, searchr1, mmsearchr1}. This decoupled optimization process not only reduces risk but also enhances transparency and controllability.

\section{Methodology}
\subsection{Overall Framework}
The overall architecture adopts a modular design and contains four main components. 1. Multimodal pathology knowledge base: This is a specialized vector database containing a rich collection of pathology textbook pages. It acts as an external storage for agent queries to collect relevant evidence, such as images of similar cases and their corresponding diagnostic descriptions; 2. Intelligent agentic router: This is the central processing unit of our framework. It accepts the initial diagnosis query, decomposes it into logically sequential subtasks, and plans; 3. VRAG Agent~\cite{vragrl}: This module supports multi-turn retrieval and summarization. It interacts with the knowledge base and distills the returned textbook images into concise, useful information. 4. Core vision language model (inference engine)~\cite{pathor1}: We use the pretrained pathology VLM as the basic inference engine. With the contextual summaries provided by the VRAG agent, it performs inference to address the diagnostic query.

\subsection{Construction of a Multimodal Pathology Knowledge Base}
To support retrieval-augmented reasoning in pathology, we construct a high-quality multimodal knowledge base that integrates authoritative textual and visual information. We curated a large corpus by collecting over 600 authoritative pathology textbooks—approximately 300,000 pages in total and, after removing irrelevant content, retained more than 200,000 high-quality pages, which were converted into image-based samples for diagnostic relevance. Using the ColQwen2 model~\cite{FaysseSWOVHC25}, we embed image-text pairs into a unified vector space that captures both visual and semantic signals. The embeddings are indexed with the HNSW algorithm~\cite{malkov2018efficient} and stored in Milvus~\cite{milvus} to support efficient high-dimensional retrieval during reasoning. Full construction details are provided in \Cref{appendix:kb_construction}.

\begin{figure*}
    \centering
    \includegraphics[width=1\linewidth]{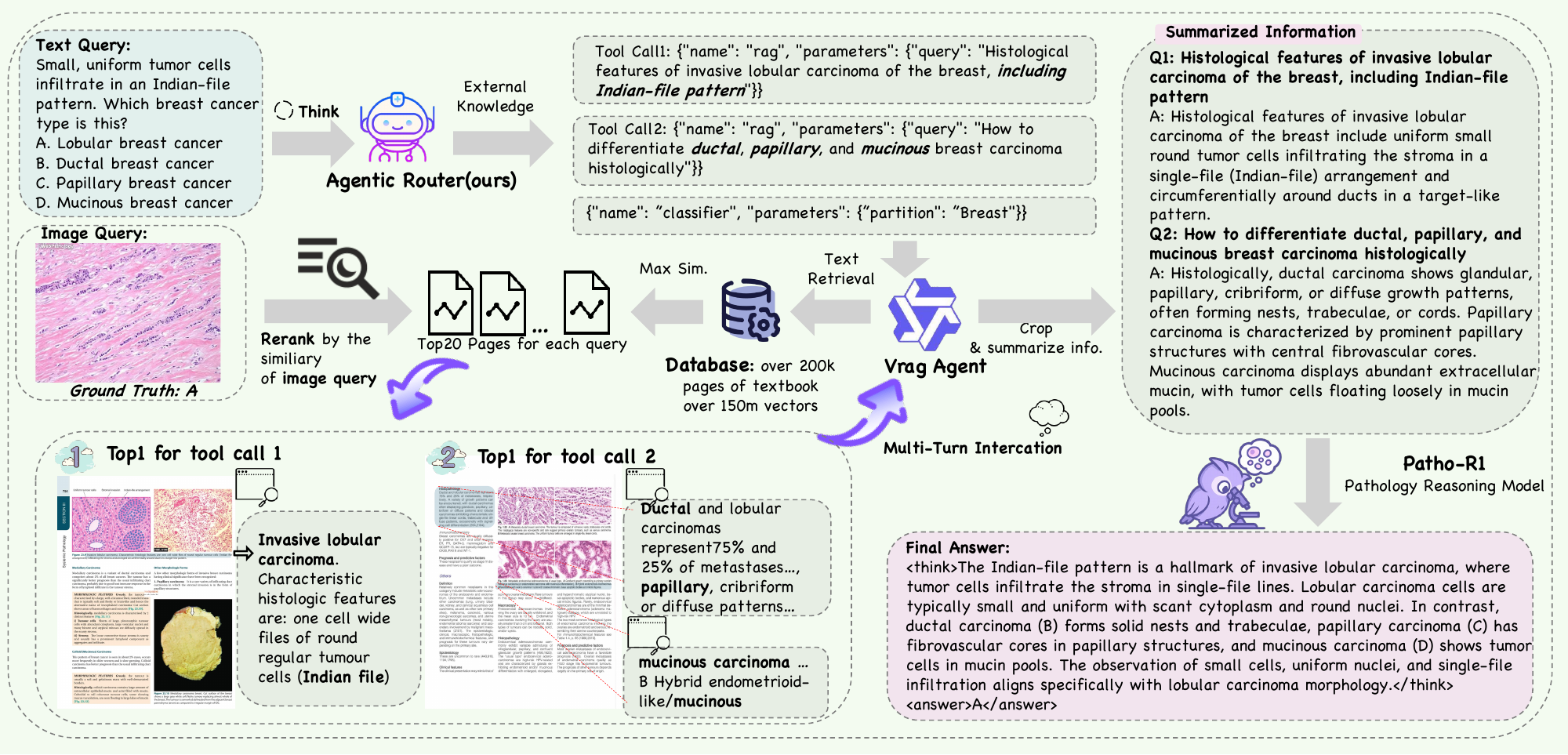}
    \caption{An illustration of the multi-turn retrieval and summarization process.}
    \label{fig: multi-turn retrieval}
\end{figure*}

    \subsection{Multimodal Fusion}
\begin{figure}
    \includegraphics[width=\columnwidth]{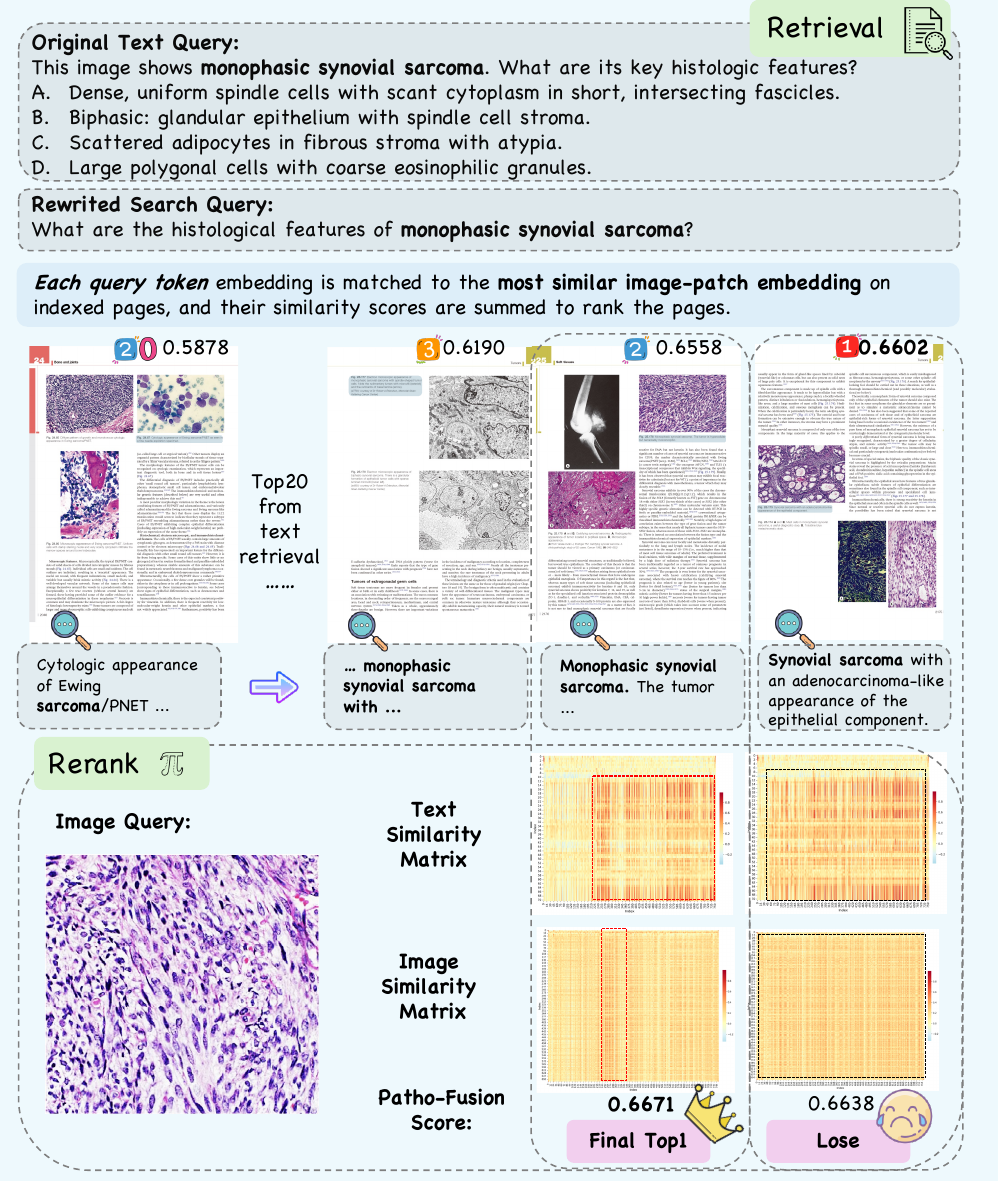}
    \caption{Illustration of the reranking process utilizing modality fusion.}
    \label{fig:fusion}
\end{figure}
    \paragraph{Method}
    Let \(S_t \in \mathbb{R}^{N_t \times N_d}\) denote the text–document similarity matrix and \(S_v \in \mathbb{R}^{N_v \times N_d}\) the image–document similarity matrix, where \(N_t\) is the number of text tokens, \(N_v\) the number of image patches, and \(N_d\) the number of document tokens. These two modalities are fused by the following expression:
    \begin{equation}
        \begin{aligned}
            \text{std}\bigl(\text{std}(S_{t}[i,:])\bigr)
            &\times \bigl[\text{mean}\bigl(\kappa(S_{t}[i,:])\bigr)\bigr]^2
            \times \text{mean}\bigl(\kappa(S_{v}[i,:])\bigr) \\
            &+\; \text{mean}\!\bigl(\max(S_{t}[i,:])\bigr),
        \end{aligned}
        \label{eq: fusion}
    \end{equation}
    where $\kappa(S_{t}[i,:])$ denotes the kurtosis of the similarity scores between all document tokens and text token $i$, and $\kappa(S_{v}[i,:])$ denotes the kurtosis of the similarity scores between all document tokens and image patch $i$. The first term captures how the standard deviation $\text{std}(\cdot)$ and kurtosis $\kappa(\cdot)$ reflect the variation of similarity scores across tokens or image patches with respect to the database document, represented as rows in $S_t$ or $S_v$. This encourages the retrieval results to have different importance for various tokens in the document, indexed by $j = 1, \ldots, N_d$. The reason for this is, in practice, only a portion of a page is typically relevant to the database document, meaning only some $j$-th elements in $S_t[i,j]$ or $S_v[i,j]$ contribute meaningfully. When all parts of a document exhibit high responsiveness, resulting in a low standard deviation, this may indicate a noisy document with problematic embeddings, which should be deprioritized. The second term, \(\text{mean}\!\bigl(\max(S_{t}[i,:])\bigr)\), as in CoPaLi~\cite{FaysseSWOVHC25}, quantifies the maximum relevance $\max(S_{t}[i,:])$ of each token to any token in the document. 
    
    It is important to note that the similarity matrix of the image modality with respect to the document embedding, i.e., $S_v \in \mathbb{R}^{N_v \times N_d}$, is used only to calculate the kurtosis $\kappa(S_{v}[i,:])$, and the average similarity information from the image modality, such as $\text{mean}(S_v[i,:])$, is not incorporated. This design emphasizes attention to the most relevant portions of a page, identified via large values of $\kappa(S_v[i,:])$. When a token or patch shows high similarity to all parts of a document, it is treated as a noisy retrieval result and is assigned a lower score during re-ranking.

\paragraph{Explanation} To intuitively understand how \Cref{eq: fusion} influences the re-ranking process, we present the similarity matrices of the first- and second-ranked documents with respect to the query text embedding and image embedding in \Cref{fig:fusion}. As shown in \Cref{fig:fusion}, the second-ranked document exhibits a more uniform similarity, or attention, across the tokens of the document embedding. In contrast, the first-ranked document demonstrates more concentrated and higher similarity on fewer tokens of the document embedding. In this case, the high similarity of the first-ranked document to the query is meaningful, as it is caused by focused attention on the most informative tokens. On the other hand, the high similarity of the second-ranked document to the query results from more diffuse attention, likely caused by noise. Thus, although the second-ranked feature initially has the highest text similarity in the retrieval process, after re-ranking by the fusion formula, it drops to second place, while the original second-ranked document is promoted to the top rank due to its combination of high similarity and concentrated attention.

\subsection{Agentic Diagnostic Workflow with Multimodal Evidence Tracing}

Our system adopts an intelligent multi-agent workflow that transforms a diagnostic query into an evidence-grounded conclusion. Given a user input (e.g., a pathology image and candidate diagnoses), the \textbf{Agentic Router} module first parses the query, decomposes it into sub-tasks aligned with each diagnostic candidate, and formulates a high-level retrieval plan. It delegates retrieval and evidence aggregation to the \textbf{VRAG Agent}, which operates under its guidance.

The \textbf{VRAG Agent} executes a multi-turn retrieval process against the multimodal knowledge base. It first conducts text-based retrieval using candidate-specific keywords, and then re-ranks the retrieved entries by evaluating image-text similarity with the query. Through iterative evidence refinement and summarization, as illustrated in \Cref{fig: multi-turn retrieval}, the agent constructs a structured prompt containing the top 1-ranked visual evidence for each candidate. This prompt is then passed to a specialized vision-language model, which performs contrastive reasoning to produce a diagnosis and evidence-grounded report. Details of each module and step are provided in \Cref{appendix:agentic_workflow}.

\subsection{Tool Integrated RL for Agentic Router}
Traditional RAG systems are often treated as static pipelines, without adapting their behavior to each query’s complexity. To overcome this limitation, we introduce a reinforcement learning (RL) framework that enables an agent to learn dynamic invocation and routing strategies~\cite{toolrl}. The agent’s task is to generate a decision path specifying whether and how to call the RAG system. Formally, given an input query $Q_{\text{orig}}$, the agent’s policy $\pi$ outputs a path $P$, optimized to maximize the expected similarity to a ground-truth decision path $P_{\text{gt}}$
\begin{equation}\max_{\pi}\mathbb{E}_{P\sim\pi(Q_{\mathrm{orig}})}\left[R_{\mathrm{final}}(P,P_{\mathrm{gt}})\right]\end{equation}
The hierarchical reward $R_{\text{final}}$ compares the generated path to the target path step by step (see \Cref{alg: Hierarchical Reward Computation}).
The agent performs a sequence of decisions to construct the path:
\begin{itemize}
    \item \textbf{Decision 1: Whether to invoke RAG?}
    \begin{itemize}
        \item \textbf{Path A (No RAG Invocation):} If the agent decides \texttt{False}, the decision process terminates. This is for simple queries that can be answered without external knowledge.  
        Final path: $\{\texttt{rag}: \texttt{False}\}$
        \item \textbf{Path B (Invoke RAG):} If \texttt{True}, proceed to the next decision.
    \end{itemize}
    
    \item \textbf{Decision 2: How to decompose the task?}  
    (Only applies if RAG is invoked)\\
    The agent may choose to rewrite the query one or more times to better surface its core semantics and improve retrieval quality. This is not a mechanical rewriting process, but a targeted transformation to better align with the retrieval engine.
    
    \item \textbf{Decision 3: Whether to use a tissue-specific classifier?}  
    (Only applies if RAG is invoked)\\
    The agent decides whether to enable a classifier to restrict retrieval to a relevant knowledge partition.
    \begin{itemize}
        \item \textbf{Path B.1 (Global Retrieval):} If \texttt{False}, RAG retrieves documents from the full corpus.  
        Final path: $\{\texttt{rag}: \texttt{True}, \texttt{rewrite\_count}: n, \texttt{classifier}: \texttt{False}\}$
        \item \textbf{Path B.2 (Classifier-Based Retrieval):} If \texttt{True}, proceed to the next decision.
    \end{itemize}
    
    \item \textbf{Decision 4: Whether the classifier assigns the query to the correct partition?}  
    (Only applies if a classifier is enabled)\\
    The agent selects a classifier from the available set $\{C_1, \dots, C_m\}$ and attempts to assign the query to the correct partition. The effectiveness of this decision depends on both the selection of the classifier and the correctness of the classification.  
    Final path: $\{\texttt{rag}: \texttt{True}, \texttt{rewrite\_count}: n, \texttt{classifier}: \texttt{True}, \texttt{partition}: C_j\}$
\end{itemize}

\begin{algorithm}[!t]
\caption{Hierarchical Reward Computation}
\label{alg: Hierarchical Reward Computation}
\begin{algorithmic}[1]
\Require Agent path $P$, Ground Truth $P_{\text{gt}}$
\State $R_{\text{final}} \gets 0$
\If{$P.\texttt{rag} \neq P_{\text{gt}}.\texttt{rag}$}
    \State \Return $R_{\text{final}}$  \Comment{Incorrect Decision 1}
\EndIf
\If{$P_{\text{gt}}.\texttt{rag} = \texttt{False}$}
    \State $R_{\text{final}} \gets 4$  \Comment{Correct Decision 1 (Path A)}
\Else
    \State $R_{\text{final}} \gets 1$  \Comment{Correct Decision 1 (Path B)}
    \If{$P.\texttt{rewrite\_count} = P_{\text{gt}}.\texttt{rewrite\_count}$}
        \State $R_{\text{final}} \gets R_{\text{final}} + 1$ \Comment{Correct Decision 2}
    \EndIf
    \If{$P.\texttt{classifier} = P_{\text{gt}}.\texttt{classifier}$}
        \If{$P.\texttt{classifier} = \texttt{False}$}
            \State $R_{\text{final}} \gets R_{\text{final}} + 2$ \Comment{Correct Decision 3 (Path B.1)}
        \Else
            \State $R_{\text{final}} \gets R_{\text{final}} + 1$ \Comment{Correct Decision 3 (Path B.2)}
            \If{$P.\texttt{partition} = P_{\text{gt}}.\texttt{partition}$}
                \State $R_{\text{final}} \gets R_{\text{final}} + 1$ \Comment{Correct Decision 4}
            \EndIf
        \EndIf
    \EndIf
\EndIf
\State \Return $R_{\text{final}}$
\end{algorithmic}
\end{algorithm}

\paragraph{RL Training with GRPO}
We use the GRPO algorithm~\cite{shao2024deepseekmath} to train the policy $\pi_\theta$. For each query $Q_{\text{orig}}$, the agent generates multiple decision paths:

\begin{equation}
G_Q = \{(P_1, r_1), (P_2, r_2), \dots, (P_n, r_n)\}
\end{equation}
where each $P_i$ is a complete decision path and $r_i$ is its corresponding reward score $r_i \in [0,4]$ based on the hierarchical reward function. We normalize the rewards within the group to compute the advantage function $A_i(P_i \mid Q)$:

\begin{equation}
A_i(P_i \mid Q) = \frac{r_i - \mu_Q}{\sigma_Q + \eta}
\end{equation}
where $\mu_Q$ and $\sigma_Q$ are the mean and standard deviation of rewards within group $G_Q$, and $\eta$ is a small constant for numerical stability.

To update the policy, we apply the GRPO objective, which extends PPO by group-wise normalized advantages and KL regularization with a reference model. Specifically,  for each group of outputs \( \{o_i\}_{i=1}^G \), from the same query, we optimize:

\begin{align}
&\mathcal{J}_{\text{GRPO}}(\theta) = 
\mathbb{E}_{q \sim P(Q), \{o_i\}_{i=1}^G \sim \pi_{\theta_{\text{old}}}(O \mid q)} \biggl[ 
\frac{1}{G} \sum_{i=1}^G \frac{1}{|o_i|} \sum_{t=1}^{|o_i|} \nonumber\\
& \quad\min \biggl( 
\frac{\pi_\theta(o_{i,t} \mid q, o_{i,<t})}{\pi_{\theta_{\text{old}}}(o_{i,t} \mid q, o_{i,<t})} \hat{A}_{i,t}, \nonumber\\
&\quad \text{clip} \left( \frac{\pi_\theta(o_{i,t} \mid q, o_{i,<t})}{\pi_{\theta_{\text{old}}}(o_{i,t} \mid q, o_{i,<t})}, 1 - \epsilon, 1 + \epsilon \right) \hat{A}_{i,t} \biggr) \nonumber\\
&\quad - \beta D_{\text{KL}}\biggl( \pi_\theta(o_{i,t} \mid q, o_{i,<t}) \!\biggm\| \!\pi_{\text{ref}}(o_{i,t} \mid q, o_{i,<t}) \biggr)
\biggr]
\end{align}
Here, \( \hat{A}_{i,t} \) is the advantage at step $t$ within each output, computed relative to other outputs for the same query. This group-wise comparison helps the agent learn from relative improvements, leading to more effective decision-making in complex reasoning paths. See \Cref{appendix:training_details} for details.

    \begin{table*}[!ht]
        \centering
        \begin{tabular}{lrrrrrrrrr}
            \hline
            Method   & Rec@1              & Rec@5              & MRR@1              & MRR@5              & MRR@20             & NDCG@1             & NDCG@5             & NDCG@20            \\
            \hline
            CoPaLi (Text)     & 0.640 (2)          & \textbf{0.900 (1)} & 0.640 (2)          & 0.734 (2)          & 0.736 (2)          & 0.740 (2)          & 0.804 (2) & 0.796 (2) \\
            CoPaLi (Image)    & 0.060 (3)          & 0.220 (3)          & 0.060 (3)          & 0.112 (3)          & 0.170 (3)          & 0.080 (3)          & 0.174 (3)          & 0.359 (3)          \\
            WeiMoCIR & 0.060 (3)          & 0.200 (4)          & 0.060 (3)          & 0.102 (4)          & 0.158 (4)          & 0.080 (3)          & 0.163 (4)          & 0.342 (4)          \\
            \textbf{Patho-Fusion (ours)}   & \textbf{0.720 (1)} & 0.880 (2)          & \textbf{0.720 (1)} & \textbf{0.777 (1)} & \textbf{0.784 (1)} & \textbf{0.820 (1)} & \textbf{0.824 (1)} & \textbf{0.827 (1)} \\
            \hline
        \end{tabular}
        \caption{Comparison of retrieval methods. Numbers in parentheses denote rank.}
        \label{tab:similarity_metrics}
    \end{table*}
\section{Experiments}

 \subsection{Multimodal Fusion}
    \paragraph{Baseline Methods}
    For baseline comparisons, we evaluate the proposed method against CoPaLi~\cite{FaysseSWOVHC25} and Weighted Modality Fusion and Similarity for Composed Image Retrieval (WeiMoCIR)~\cite{wu2024training}.
    \begin{itemize}
        \item For CoPaLi~\cite{FaysseSWOVHC25}, which supports only a single modality, we apply retrieval separately for each modality. The scoring function is
        \begin{equation}
            \sum_{i=1}^{N_q}\max_{j=1,\ldots,N_d}\bigl\langle E_q(i),\,E_d(j)\bigr\rangle,
        \end{equation}
        where \(E_q\) is either \(E_t\) or \(E_v\) so \(N_q=N_t\) or \(N_q=N_v\), and \(\langle\cdot,\cdot\rangle\) denotes the inner product. Each query embedding $E_q(i)$ is matched to the most relevant document embedding $E_d(j)$, where $j = 1, \ldots, N_d$, and the results are aggregated over all query tokens or patches.
        \item In WeiMoCIR, the query embedding is computed as
        \begin{equation}
            \mathbf{q} = (1 - \alpha) \cdot \mathbf{e}_v + \alpha \cdot \mathbf{e}_t,
        \end{equation}
        where \(\mathbf{e}_v = \text{mean}(E_v(i,:))\) is the average vision embedding over all patches \(i = 1, \ldots, N_v\), and \(\mathbf{e}_t = \text{mean}(E_t(i,:))\) is the average text embedding over all tokens \(i = 1, \ldots, N_t\). The parameter \(\alpha = 0.1\) represents the weighting coefficient for the text modality. The final similarity score between the query and a database document is computed using the inner product as
        \begin{equation}
            \frac{1}{N_d}\sum_{j=1}^{N_d} \left\langle \mathbf{q},\,\mathbf{e}_{d,j} \right\rangle,
        \end{equation}
        where \(\mathbf{e}_{d,j}\) is the \(j\)-th token embedding of the document.
    \end{itemize}

    \paragraph{Dataset and Evaluation Protocol} The dataset consists of 100 pairs of images, questions, and answers curated by domain experts. We randomly split the dataset, using 50\% for training and 50\% for testing. The modality fusion function is optimized only on the training data to prevent potential data leakage. All modality fusion methods are evaluated on the test set, and recall, mean reciprocal rank (MRR), and normalized discounted cumulative gain (NDCG) metrics are reported.

    \paragraph{Experimental Results}
    The experimental results are shown in \Cref{tab:similarity_metrics}. Recall@20 is omitted since there are only 20 results in the re-ranking stage and the Recall@20 is identical for all algorithms. The results demonstrate a clear advantage for the proposed modality fusion method over the baseline approaches. While using only the text modality for retrieval can already achieve good performance, the retrieval results remain suboptimal without the proposed fusion strategy in \Cref{eq: fusion}. Methods using only the image modality or WeiMoCIR perform worse than the proposed fusion by a significant margin. This is primarily because both heavily rely on the image modality for retrieval, whereas pathology images require strong domain expertise to interpret and general-purpose embedding methods may not provide optimal representations for this task. Although WeiMoCIR achieves strong results on general retrieval benchmarks, it underperforms in the medical multimodal retrieval setting. Overall, these findings demonstrate that general multimodal fusion strategies are not sufficient for the pathology domain and the specifically designed fusion mechanism proposed here offers superior effectiveness.

\begin{figure*}
    \centering
    \includegraphics[width=1\linewidth]{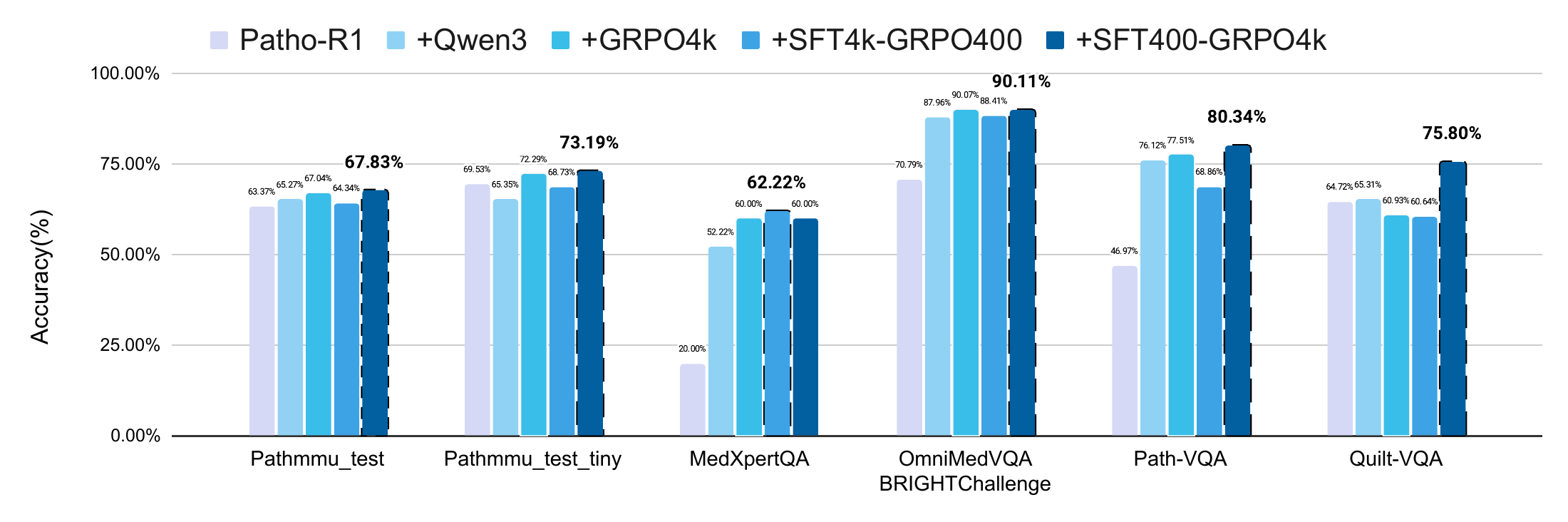}
    \caption{Ablation study results across multiple medical QA datasets.}
    \label{fig:ablation}
\end{figure*}

\begin{table*}[!ht]
\centering
\renewcommand{\arraystretch}{1}
\setlength{\tabcolsep}{2pt} 
\begin{tabular}{lccccccccccc}
\hline
\multirow{2}{*}{\textbf{Model}} & \multicolumn{5}{c}{\textbf{PathMMU-test}} & \multicolumn{5}{c}{\textbf{PathMMU-test-tiny}} \\
\cline{2-11}
 & Atlas & EduContent & PathCLS & PubMed & SocialPath & Atlas & EduContent & PathCLS & PubMed & SocialPath \\
\hline
InternVL2-8B & 43.68 & 44.86 & 23.77 & 44.56 & 45.40 & 46.63 & 50.59 & 21.47 & 49.11 & 51.38 \\
InternVL2.5-8B & 50.06 & 50.62 & 32.84 & 50.02 & 50.87 & 51.44 & 50.59 & 29.38 & 55.87 & 57.80 \\
InternVL3-8B & 54.07 & 50.80 & 39.09 & 54.04 & 53.32 & 58.17 & 54.90 & 42.94 & 57.65 & 60.55 \\
Llama-3.2-11B-VI & 41.05 & 37.49 & 26.72 & 38.82 & 39.21 & 45.19 & 38.04 & 29.38 & 39.50 & 41.74 \\
Llama-3.2V-11B-cot & 51.81 & 45.45 & 30.76 & 48.15 & 46.10 & 49.04 & 47.06 & 29.94 & 53.38 & 45.41 \\
LLaVA-Onevision-7B & 21.65 & 21.27 & 12.01 & 27.77 & 21.25 & 31.25 & 21.18 & 13.56 & 31.32 & 18.35 \\
Qwen2.5VL-7B & 41.18 & 43.20 & 24.82 & 42.77 & 39.67 & 44.23 & 49.41 & 24.86 & 44.84 & 40.83 \\
\textbf{Patho-R1-7B} & \underline{75.34} & \underline{66.43} & \underline{45.40} & \underline{66.06} & \underline{67.93} & \textbf{81.73} & \underline{75.29} & \underline{44.63} & \underline{72.24} & \underline{67.89} \\
\textbf{Patho-AgenticRAG} & \textbf{78.32} & \textbf{70.96} & \textbf{53.16} & \textbf{69.69} & \textbf{71.06} & \underline{79.33} & \textbf{76.47} & \textbf{57.22} & \textbf{72.24} & \textbf{74.70} \\
\hline
\end{tabular}
\captionsetup{justification=raggedright,singlelinecheck=false}
\caption{Comparison of model performance across multiple tasks. The left group shows results on PathMMU-test, and the right group on PathMMU-test-tiny. Best and second-best performances are bolded and underlined respectively.}
\label{mcq1}
\end{table*}

\begin{table}[!ht]
\centering
\renewcommand{\arraystretch}{1}
\setlength{\tabcolsep}{2pt} 
\begin{tabular}{lcccc}
\hline
\multirow{2}{*}{\textbf{Model}} & \multicolumn{2}{c}{\textbf{YorN}} & \textbf{MedXpert} & \textbf{OmniMed} \\
\cline{2-5}
  & Quilt & Path & Path & Bright \\
\hline
InternVL2-8B & 60.56 & 61.36 & 10.00 & 40.56 \\
InternVL2.5-8B & 60.06 & \underline{64.78} & \underline{22.22} & 49.78 \\
InternVL3-8B & 33.82 & 18.56 & 15.56 & 65.28 \\
Llama-3.2-11B-VI & 63.27 & 63.50 & 13.33 & 47.08 \\
Llama-3.2V-11B-cot & 54.81 & 56.42 & 21.11 & 54.83 \\
LLaVA-Onevision-7B & 24.20 & 52.38 & 16.67 & 31.46 \\
Qwen2.5VL-7B & 52.19 & 41.82 & 12.22 & 43.60 \\
\textbf{Patho-R1-7B} & \underline{64.72} & 46.97 & 22.00 & \underline{70.79} \\
\textbf{Patho-AgenticRAG} & \textbf{75.80} & \textbf{80.34} & \textbf{60.00} & \textbf{90.11} \\
\hline
\end{tabular}
\captionsetup{justification=raggedright,singlelinecheck=false}
\caption{Performance comparison on Quilt-VQA, Path-VQA, MedXpert, and OmniMed.}
\label{mcq1_medonly}
\end{table}

\subsection{Patho-AgenticRAG Evaluation Results}

\paragraph{Ablation Analysis}
We conducted three main ablation studies to investigate the necessity and data proportion of SFT before GRPO. The results show that skipping SFT leads to poor convergence during GRPO. However, performing SFT with a large amount of data causes the model to lack generalizable capabilities and exhibit rigid output patterns. Therefore, using a small amount of SFT data as a cold start before GRPO is the optimal strategy. Notably, adopting a lightweight SFT phase (e.g., SFT400) before GRPO achieves the best overall balance. This setting consistently outperforms both the "no-SFT" and "large-SFT" baselines across multiple datasets. For example, on the Path-VQA benchmark, using SFT400+GRPO4k improves performance from 77.51\% (GRPO4k only) to 80.34\%. Similarly, on the Quilt-VQA dataset, performance improves from 60.93\% (GRPO4k) to 75.80\%, a +14.87\% increase, indicating that a small amount of supervised guidance before preference optimization significantly enhances model capability. These results suggest that SFT400 provides an effective “cold start” that guides the policy initialization without compromising flexibility or generalization, as shown by Figure~\ref{fig:ablation}.
\paragraph{Close-Ended Benchmarks Results}

Closed-ended questions play a crucial role in pathology-related tasks, particularly in diagnostic classification. To evaluate model performance on such tasks, we consider two types of close-ended question datasets: (1) Yes/No questions, selected from Path-VQA~\cite{pathvqa} and Quilt-VQA~\cite{quilt1m}; and (2) multiple-choice questions, sourced from PathMMU \cite{pathmmu}, MedXpertQA \cite{medxpertqa}, and OmniMedVQA \cite{omnimedvqa}.

The results on close-ended benchmarks are summarized in \Cref{mcq1,mcq1_medonly}. Patho-AgenticRAG achieves the best overall performance across most tasks, significantly outperforming both general-purpose vision-language models (e.g., InternVL3, Qwen2.5VL) and domain-specialized baselines such as Patho-R1-7B~\cite{pathor1}. Specifically, Patho-AgenticRAG achieves +13.37\% improvement on Quilt-VQA (75.80\% vs. 64.72\%) and +38.00\% on MedXpertQA (60.00\% vs. 22.00\%) over Patho-R1. The largest margin appears on MedXpertQA, highlighting the importance of retrieval-augmented reasoning in knowledge-intensive tasks. On OmniMedVQA Bright Challenge, the model improves from 70.79\% (Patho-R1) to 90.11\%, a +19.32\% increase, demonstrating substantial gains in both generalization and diagnostic precision. Details are in \Cref{appendix:experimental_prompts}

\section{Conclusion}
We proposed Patho-AgenticRAG, a novel multimodal retrieval-augmented generation framework tailored for pathology diagnosis. By leveraging intelligent agents for dynamic querying of image-based vector databases, as well as employing task decomposition, query planning, and evidence aggregation, our approach significantly enhances the reasoning capabilities of vision-language models in pathology tasks.
Our framework addresses the critical issue of hallucination in pathology diagnosis by promoting knowledge alignment, supporting evidence-based reasoning, and improving factual consistency in generated outputs. Patho-AgenticRAG demonstrates significant improvements over existing state-of-the-art multimodal models in key metrics, including answer precision and evidence traceability, representing a notable advancement in the integration of image content and reasoning for real-world pathology applications.

\bibliography{aaai2026}

\clearpage
\onecolumn
\appendix
\section{Multimodal Knowledge Base Construction}
\label{appendix:kb_construction}
To construct a high-quality knowledge base tailored for pathology, we collected and processed over 600 authoritative pathology textbooks that span a wide range of diagnostic domains. These sources were chosen for their depth, credibility, and relevance to real-world diagnostic practice.

To improve retrieval accuracy and reduce noise in downstream reasoning, we removed sections that typically lack diagnostic content, including covers, prefaces, tables of contents, and references. The remaining pages, primarily rich in textual explanations and pathology figures, were preserved as the core knowledge units. These preserved pages were further categorized into 19 distinct anatomical and diagnostic classes to support systematized retrieval. Specifically, the categories are as follows:
\begin{enumerate}[leftmargin=2.5em]
    \item \textbf{Bone and soft tissue} – including neoplastic and non-neoplastic lesions of bone, cartilage, and connective tissue.

    \item \textbf{Cytology} – encompassing exfoliative and fine-needle aspiration cytopathology across body systems.

    \item \textbf{Gastrointestinal tract, liver, gallbladder, pancreas, and digestive system} – covering both luminal and solid organ pathology within the digestive system.

    \item \textbf{Hematology, lymphatic system, and bone marrow} – focusing on hematopoietic and lymphoid neoplasms and marrow disorders.

    \item \textbf{Infectious diseases} – including pathology of viral, bacterial, fungal, and parasitic infections.

    \item \textbf{Oral and head and neck} – covering salivary glands, oral mucosal lesions, larynx, pharynx, nasal cavity, and other ENT (ear, nose, throat) structures.

    \item \textbf{Urinary and male reproductive system} – involving kidney, bladder, prostate, testis, and related structures.

    \item \textbf{Breast} – dedicated to benign, precursor, and malignant breast lesions.

    \item \textbf{Endocrine system} – covering thyroid, parathyroid, adrenal glands, and neuroendocrine tumors.

    \item \textbf{General comprehensive pathology} – consisting of cross-system concepts such as inflammation, necrosis, and neoplasia.

    \item \textbf{Histology and embryology} – including normal tissue architecture and developmental processes.

    \item \textbf{Neonatal, pediatric, and childhood diseases} – covering age-specific pathologies from birth to adolescence.

    \item \textbf{Skin and dermatopathology} – addressing inflammatory, infectious, and neoplastic skin conditions.

    \item \textbf{Central nervous system} – including brain, spinal cord, and peripheral nerve pathology.

    \item \textbf{Female reproductive system} – encompassing ovary, uterus, cervix, and vulvovaginal tract pathology.

    \item \textbf{Gross specimen sampling} – providing guidance on macroscopic examination and tissue sectioning.

    \item \textbf{Immunohistochemistry and molecular pathology} – focused on biomarker detection and molecular diagnostics.

    \item \textbf{Ophthalmology and otolaryngology} – covering diseases of the eye, ear, and related sensory organs.

    \item \textbf{Trachea, lung, pleura, respiratory system, and mediastinum} – including neoplastic and inflammatory diseases of the thoracic cavity.
\end{enumerate}

Each cleaned PDF was converted page-wise into an image format, ensuring that the visual structure (including images, captions, and layout) remained intact.

We used the ColQwen2 model~\cite{FaysseSWOVHC25}, a powerful vision-language encoder, to generate unified high-dimensional embeddings for each image-text pair. This model jointly encodes both the fine-grained visual features and the contextual diagnostic semantics into a shared vector space, enabling effective multimodal similarity retrieval.

All embeddings were stored in Milvus~\cite{milvus}, an open-source vector database optimized for large-scale similarity search. To accelerate retrieval in high-dimensional space, we employed the hierarchical navigable small world (HNSW) indexing algorithm~\cite{malkov2018efficient}, which is well suited for fast, approximate nearest neighbor search at scale. This infrastructure allows our reasoning agents to retrieve highly relevant visual evidence from hundreds of thousands of multimodal entries in real time.

\section{Details of the Agentic Diagnostic Workflow}
\label{appendix:agentic_workflow}

The intelligent diagnostic pipeline operates in four distinct stages, involving three key modules: the Agentic Router (task planner), the VRAG Agent (retrieval and summarization), and the Pathology VLM (inference engine). Each step is described in detail below.

\textbf{Step 1: Query Understanding and Task Decomposition (Agentic Router)}  
The Agentic Router receives the user’s input and identifies the semantic structure of the task, typically involving image understanding and candidate differentiation. It decomposes the query into subproblems mapped to each candidate diagnosis.

\textbf{Step 2: Entity-Centric Multimodal Retrieval (within VRAG Agent)}  
For each candidate diagnosis, the VRAG Agent performs a two-step retrieval process:
\begin{itemize}
    \item \textit{Textual Retrieval:} Using candidate-specific keywords, the system retrieves the top 20 textual entries from the knowledge base.
    \item \textit{Multimodal Re-ranking:} Retrieved entries are then ranked by calculating joint similarity between the input image and associated image-text pairs using the ColQwen2 embedding~\cite{FaysseSWOVHC25}. The similarity score is computed according to Equation 1, which captures the cross-modal relevance between the query image and each retrieved image-text pair in the embedding space. The highest-ranked entry becomes the evidence for that candidate.
\end{itemize}

\textbf{Step 3: Iterative Evidence Aggregation (VRAG Agent)}  
The VRAG Agent~\cite{vragrl} performs multi-turn retrieval and refinement. It dynamically evaluates evidence sufficiency, performs focused re-retrieval when necessary, and aggregates the most discriminative content into a structured prompt. The prompt includes: (1) the original user input, (2) top retrieved evidence for each diagnosis, and (3) explicit comparative instructions.

\textbf{Step 4: Reasoning and Report Generation (Pathology VLM)}  
The structured prompt is passed to a specialized vision-language model fine-tuned for diagnostic tasks~\cite{pathor1}. The model conducts contrastive reasoning, constrained to the provided evidence, and returns both a predicted diagnosis and an interpretable justification.

\section{Details of Models and Training}
\label{appendix:training_details}
\subsection{Training Dataset Construction}

\begin{figure}[h]
    \centering
    \includegraphics[width=1\linewidth]{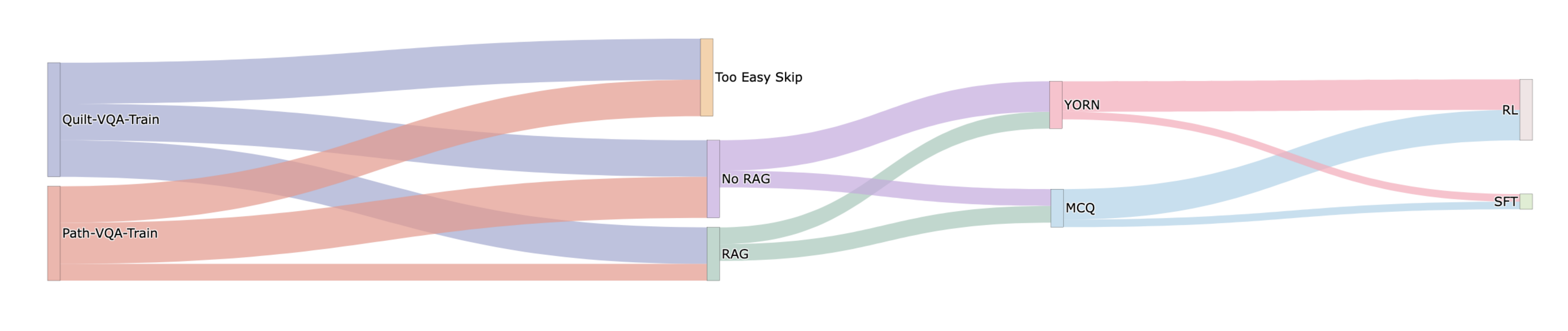}
    \caption{Training Data Composition for Agentic Router}
    \label{fig:train_data}
\end{figure}

The training data for the agentic router was derived from the training sets of Quilt-VQA\cite{quilt1m} and Path-VQA\cite{pathvqa}. We first ran inference using Patho-R1 and filtered the samples based on the correctness of its responses. Specifically, we removed questions with overly simple reasoning or insufficient thought chains, as they were not useful for training our agent flow. From the remaining samples, we randomly selected 2,200 questions that Patho-R1 answered incorrectly (indicating a need for RAG) and 2,200 that it answered correctly (indicating no need for RAG) to construct the training dataset. Among these selected questions, we ensured an even split between MCQ and YORN formats. We then distilled the ground truth using QwenMax, ultimately retaining 4,000 samples for RL training and an additional 400 samples for SFT training, as shown in \Cref{fig:train_data}.

Specifically, to construct ground-truth decision paths \( P_{\text{gt}} \) for reinforcement learning, we prompted \texttt{QwenMax} with carefully designed instructions that emulate expert diagnostic reasoning. Specifically, given a user query \( Q_{\text{orig}} \), we provided \texttt{QwenMax} with a prompt that specifies: (1) available tools (e.g., \texttt{rag}, \texttt{classifier}), (2) a comprehensive list of anatomical partitions, and (3) detailed behavioral guidelines for tool invocation.

The model was instructed to first output a \texttt{<think>} field capturing its reasoning, followed by one or more structured \texttt{<tool\_call>} entries. For instance, it would decide whether to query external knowledge via the \texttt{rag} tool, invoke the \texttt{classifier} tool to infer tissue system based on question content, or rely solely on internal knowledge when appropriate. A sample prompt template and tool call schema are provided in Appendix~\ref{appendix:experimental_prompts}.

The resulting outputs serve as high-quality ground-truth action trajectories, reflecting both \textit{whether} to invoke the retrieval system and \textit{how} to structure the tool usage. These trajectories form the supervision signal for training the agent’s policy \( \pi \), which aims to recover the expert-like decision plan \( P_{\text{gt}} \) from an input query.

\subsection{Training Details}

All tool calls in the dataset are represented in JSON format. We use Qwen3-4B as the base model for all experiments, and training was conducted on 8 NVIDIA RTX 4090 GPUs.

\textbf{Supervised fine-tuning:}
We adopted the LLaMA-Factory\footnote{\url{https://github.com/hiyouga/LLaMA-Factory.git}} framework and froze the vision tower. We used a learning rate of 1e-5 and trained on 400 samples for 3 epochs. 

\textbf{Reinforcement learning:}
We adopted the verl\footnote{\url{https://github.com/volcengine/verl}} framework for reinforcement learning. We set the actor and critic learning rates 1e-6 and 1e-5 respectively. For GRPO, we trained on 4k samples for 3 epochs

\section{Experimental Benchmarks, Evaluation Metrics, and Prompts}
\label{appendix:experimental_prompts}
We presented the prompts used throughout our Patho-AgenticRAG workflow, along with the benchmarks and evaluation protocols adopted for assessment.
\subsection{Multimodal Benchmarks for Pathology}
We evaluate our approach using multiple multimodal benchmarks relevant to pathology. For the multiple-choice (MCQ) setting, we adopt the PathMMU dataset, a pathology-specific benchmark comprising images and expert-annotated questions. Following the dataset’s protocol, we downloaded pathology images originally shared via Twitter. However, due to post deletions, some images were no longer accessible, and the corresponding questions were removed from evaluation.

The PathMMU-test-tiny split includes a total of 1,139 questions, distributed as follows: Atlas (208), EduContent (255), PathCLS (177), PubMed (281), and SocialPath (218). The full PathMMU-test split contains 8,454 questions: Atlas (799), EduContent (1,683), PathCLS (1,632), PubMed (2,787), and SocialPath (1,553).

To broaden evaluation coverage, we further curated pathology-focused subsets from general medical VQA datasets. Specifically, we selected 90 pathology-related examples from MedXpertQA, and used the BRIGHT Challenge subset (890 cases) from OmniMedVQA, which focuses on diagnostic reasoning across medical specialties.

Finally, for the YORN dataset, we collected closed-ended questions from the test splits of Path-VQA and Quilt-VQA, resulting in 3,362 and 343 questions, respectively.

\subsection{Evaluation Metrics for Multi-Model Fusion}
    We evaluate retrieval performance using three metrics: mean reciprocal rank at $k$ (MRR@$k$), recall@$k$, and normalized discounted cumulative gain at $k$ (NDCG@$k$), defined as follows:

    \begin{itemize}
        \item \textbf{Recall@$k$} measures the fraction of queries for which the target textbook page is included among the top $k$ retrieved results. Recall@$k$ is given by
        \begin{equation}
            \mathrm{Recall}@k = \frac{1}{|Q|} \sum_{i=1}^{|Q|} \mathbb{I}\left(\mathrm{rank}_i \leq k\right),
        \end{equation}
        where $\mathbb{I}$ is the indicator function.

        \item \textbf{Mean Reciprocal Rank at $k$ (MRR@$k$)} reports the average reciprocal rank of the target textbook page within the top $k$ results. Let $|Q|$ be the number of queries and $\mathrm{rank}_i$ the position of the target textbook page for query $i$. The metric is defined as
        \begin{equation}
            \mathrm{MRR}@k = \frac{1}{|Q|} \sum_{i=1}^{|Q|} \frac{1}{\min\left(\mathrm{rank}_i,\, k+1\right)},
        \end{equation}
        where the reciprocal rank is zero if the target textbook page does not appear in the top $k$.

        \item \textbf{Normalized Discounted Cumulative Gain at $k$ (NDCG@$k$)} evaluates ranking quality based on the graded relevance of retrieved textbook pages. For NDCG calculation, the target textbook page is assigned a relevance score of $2$, and the previous and next textbook pages are assigned a score of $1$, reflecting that neighboring pages may contain partially relevant information. NDCG@$k$ is computed as
        \begin{equation}
            \mathrm{NDCG}@k = \frac{1}{|Q|} \sum_{i=1}^{|Q|} \frac{\mathrm{DCG}_i@k}{\mathrm{IDCG}_i@k},
        \end{equation}
        where
        \begin{equation}
            \mathrm{DCG}_i@k = \sum_{j=1}^{k} \frac{2^{\mathrm{rel}_{i,j}} - 1}{\log_2(j+1)}
        \end{equation}
        and $\mathrm{rel}_{i,j}$ is the relevance score of the $j$-th retrieved textbook page for query $i$. The ideal DCG, $\mathrm{IDCG}_i@k$, is computed with pages sorted by the highest possible relevance.
    \end{itemize}

\subsection{Evaluation Metrics for Medical QA Tasks}
To evaluate model performance on various pathology benchmarks, we adopt standard accuracy-based metrics. For closed-ended visual question answering tasks, including multiple-choice and yes-or-no questions, we compute exact-match accuracy, where a prediction is considered correct only if it exactly matches the ground-truth answer. In addition, we report per-category scores on datasets such as PathMMU, whose subsets originate from diverse real-world sources (e.g., textbooks, educational slides, social media, and literature). This allows us to assess the model’s generalization ability across heterogeneous pathology domains.

\subsection{Prompts}
\begin{tcolorbox}[
  coltext=black,         
  colframe=black,        
  title= Prompt for Agentic Router,
  coltitle=white,        
  fonttitle=\bfseries, 
  fontupper=\ttfamily,
  boxrule=0.1mm,         
  arc=2mm,               
  enhanced
]
You are a helpful dialogue assistant capable of leveraging tool calls to solve user tasks and provide structured chat responses.\\

 \#\#\# User Query

\{query\}\\

 \#\#\# Available Tools
 
In your response, you can use the following tools:
\{'rag', 'classifier'\}\\

 \#\#\# Partitions:

\{
    "Bone\_and\_Soft\_Tissue", 
    "Cytology", 
    
    "Gastrointestinal\_Tract\_Liver\_Gallbladder\_Pancreas\_Digestive\_System", 
    
    "Hematology\_Lymphatic\_System\_and\_Bone\_Marrow", 
    
    "Infectious\_Diseases", "Oral\_and\_Head\_Neck",
    
    "Urinary\_and\_Male\_Reproductive\_System", 
    
    "Breast", "Endocrine", 
    "General\_Comprehensive", 
    "Histology\_and\_Embryology",
    
    "Neonatal\_Pediatric\_and\_Child",
    "Skin\_Dermatology", "Central\_Nervous\_System",
    
    "Female\_Reproductive\_System",
    "Gross\_Specimen\_Sampling",
    
    "Immunohistochemistry\_and\_Molecular\_Pathology",
    "Ophthalmology\_Otolaryngology",
    
    "Trachea\_Lung\_Pleura\_Respiratory\_System\_and\_Mediastinum"
\} \\

\#\#\# Chain of Thoughts

- You must always include the <think> field to outline your reasoning. Decide whether to use <tool\_call> (possibly multiple times).

- Given a user query, first assess whether external knowledge is needed. If so, call the rag tool.

- If the question is in multiple-choice or yes/no format, do not blindly convert each option into a question. Instead, analyze what knowledge is required to choose the correct answer — e.g., differential histological features, diagnostic criteria, cellular morphologies — and formulate targeted queries that reflect those specific informational needs.

- You may still use the options as anchors, but query construction should aim for high signal-to-noise retrieval concise, informative, and focused.

- The rag tool should receive an informative, purpose-driven query as its parameter.

- After determining that rag is needed, further inspect whether the query includes indicators of tissue classification or pathology specimen classification (e.g. mention of specific tissue, organ system, tissue structure, etc.). If so, invoke the classifier tool in parallel.

- If the category cannot be determined from the question or option, there is no need to call classifier, just call rag.

- The classifier tool must always be invoked with a partition parameter whose value is exactly one of the predefined partition names. These are case-sensitive and must come from the provided list of partitions.

- The output format for classifier must strictly follow this structure:\{ "name": "classifier", "parameters": \{ "partition": "<Exact\_Partition\_Name\_From\_List>" \}\}

- When the question and options rely mainly on an image to make a distinction, just provide a <think> field to explore possible histological interpretations based on general principles. Do not call any tools.

- Similarly, if the question reflects common knowledge in pathology or histology, such as "What stain is used for nuclei?" or "Which cell secretes collagen?", a <think> is sufficient, and no tool needs to be called.\\

 \#\#\# Output Format
 
<think> Your thoughts and reasoning </think>

<tool\_call>

\{"name": "Tool name", "parameters": \{""\}

 \{"name": "... ...", "parameters": \{"... ..."\}\}\}
 
...

</tool\_call>

\end{tcolorbox}

\begin{tcolorbox}[
  coltext=black,         
  colframe=black,        
  title= Prompt for Vrag Agent,
  coltitle=white,        
  fonttitle=\bfseries, 
  fontupper=\ttfamily,
  boxrule=0.3mm,         
  arc=2mm,               
  enhanced
]
Your task is **not** to directly answer the question, but to assist a downstream system by **retrieving and summarizing factual information** relevant to answering it.

You should focus on **describing the characteristics, criteria, or diagnostic features** associated with the possible answer—**not** on concluding what the answer is. \\

For example, if the question is whether an image shows invasive ductal carcinoma, you should **retrieve facts about how invasive ductal carcinoma typically presents**, especially in medical images or pathology reports. \\

Use reasoning steps enclosed in <think>...</think> to guide your process. \\

You may request external information using <search>...</search> or issue search queries using <search> query </search>. The user will provide relevant results. 
If a new image is retrieved, you may optionally crop it using <bbox>[x1, y1, x2, y2]</bbox>.\\

Base your reasoning on the image, the question, and any supporting data you collect. 
Use as many retrievals as necessary. Never rely solely on the initial image.

Once you have gathered enough information, provide a concise summary of the most **relevant evidence or diagnostic criteria** within <answer> and </answer> tags. \\

Do not state whether the diagnosis or conclusion is correct—just provide the features or findings that are relevant for making that decision.\\

Question: 

\{question\}
\end{tcolorbox}

\begin{tcolorbox}[
  coltext=black,         
  colframe=black,        
  title= Prompt for Patho-R1,
  coltitle=white,        
  fonttitle=\bfseries, 
  fontupper=\ttfamily,
  boxrule=0.1mm,         
  arc=2mm,               
  enhanced
]

You are a pathology expert, your task is to answer question step by step. 

Use the following format:

<think> Your step-by-step reasoning, you could use the context below if needed </think>

<answer> Your final answer, no explanation. </answer>\\

Question: 
\{query\}\\

Context: 
\{information summrized by vrag\}

\end{tcolorbox}

\end{document}